\documentclass[11pt]{article}
\usepackage{cancel}
\usepackage{eurosym}

\usepackage{hyperref}
\usepackage[ruled, lined]{algorithm2e}

\usepackage{amssymb}
\include{macros}

\usepackage{ifthen}
\newlength{\margintonumber}
\newcommand{\numline}[2]{
   \ifthenelse{#2>-1}{
     \setlength{\margintonumber}{#2\algomargin}
     \addtolength{\margintonumber}{2em}
     \hspace{-\margintonumber}
     {\textsf{\footnotesize\raggedleft\textbf{#1}}}
     \settowidth{\margintonumber}{\textsf{\footnotesize\raggedleft\textbf{#1}}}
     \addtolength{\margintonumber}{.427in}
     \hspace{-\margintonumber}
     \setlength{\margintonumber}{#2\algomargin}
     \addtolength{\margintonumber}{2em}
     \hspace{\margintonumber}
   }
   { 
     \typeout{^^JError using command \\numline (argument #2 should be
positive)^^J}
   }
   }

\begin{document}

\title{A Pebble in the AI Race}

\author{Toby Walsh \\ School of Computer Science
and Engineering, and Data61 \\ University of New South Wales,
Australia.  \\
{\tt tw@cse.unsw.edu.au }  }

\maketitle
\date

\begin{abstract}
Bhutan is sometimes described as ``a pebble between two boulders'',
a small country 
caught between the two most populous nations on earth: India and
China. This pebble is, however, about to be caught up in a vortex: the
transformation of our economic, political and social orders 
by new technologies like Artificial Intelligence. 
What can a small nation like Bhutan hope to do in the face of 
such change? What should the nation do, not just
to weather this storm, but to become a better place in which 
to live? 
\end{abstract}

\section{Introduction}

There is little doubt that Artificial Intelligence (AI)
is transforming the current economic, political and societal 
landscape. A study by PwC in 2017 estimated that
global GDP could be up to 14\% higher in
2030 as a result of AI – the equivalent of
an additional \$15.7
trillion in inflation adjusted terms \cite{pwc}.
This would likely make it the largest commercial
opportunity in today’s fast changing economy.
We also see AI disrupting political and societal
structures, from AI powered face recognition software
being used to surveil populations in China, to DeepFake videos
being used to win over voters in India \cite{deepfakes}. 

While the disruption will be great, 
change will not be evenly spread. The PwC study
estimated that the GDP of China might grow by over 25\%, 
that of the US by around 15\%, but developing
countries in Asia and elsewhere might see growth in their GDP of
just 5\%. In addition to economic growth, there might be significant 
impact on employment as AI and automation take over
more tasks. A famous (and somewhat disputed) 2013 study from the University of 
Oxford estimated that 47\% of jobs in the US were at risk of
automation \cite{frey}. In October 2016, Jim Long Kim, the President of the World
Bank, cited research predicting 69\% of jobs in India at risk, 77\% in China
and 85\% in Ethiopia \cite{frey2}. Jobs that were recently outsourced
to the developing world might, for instance, return to robots and softbots in the 
developed world. What then can a small, developing nation like Bhutan
do in the face of such immense forces? 

\section{What is AI?}

It helps to understand a little better what AI is, and where it came 
from. In this way, a better understanding might emerge of where it
might takes us. 
Artificial Intelligence started proper in 1956 when one
of its founding fathers, John McCarthy proposed the
name for the now famous Dartmouth Conference in New
Hampshire that brought together many of the founding fathers\footnote{AI has
been plagued by gender inequality since the very beginning. There are
no records of any female scientists participating in the Dartmouth meeting.   }
of the field. 
However, the history of Artificial Intelligence goes much further back than 1956
when McCarthy coined the name for the subject. It goes back before even the
invention of the computer. Humankind has been thinking
about machines that might think, and how we might
model (human) thinking for centuries. 
You can
trace back the intellectual roots of AI to the 3rd
century BC when Aristotle founded the field of formal
logic. Without logic, we would not have the modern digital
computer. And logic has often and continues to be seen
as a model for thinking, a means to make precise how we
reason and form arguments. 
You can also find echoes in the Lie Zi text,
an ancient volume of stories from China 
written sometime in the 4th century BC. 
One of the stories is of an amazing automata,
a human like ``robot''  that could sing built by Yan Shi around 1000 BC and
demonstrated to the fifth king of the Chinese Zhou Dynasty, King Mu. 

It took the invention of the digital computer in the 1950s to turn
some of these ideas into reality. 
Since then, driven by advances in computing,
Artificial Intelligence has started to make 
significant progress at automating tasks that only humans 
previously could do: perceiving the world, understanding language
and speech, reasoning about the world, and acting in that world. 
One of the misconceptions about AI is that it is one thing. Perhaps
in some distant future, we will build an all powerful AI system like
we see in Hollywood movies. But  today, AI is a collection of tools
and technologies that let computers perceive the world, learn from data, reason and act. 
And these technologies work poorly together. We can 
build AI systems that work at human or super-human level, but only
in narrow domains such as reading X-rays, playing the ancient
game of Go, or transcribing spoken Mandarin into written English. 

Another misconception is that AI is synonymous with machine learning
(ML).  In reality, machine learning is just one, albeit important, part of
AI.  Learning is an important part of our human intelligence.
Many of the intelligent skills you have today, you learnt. 
You learnt to read, to write, to do mathematics. But just as learning
is only one aspect of human intelligence, machine learning is only
one aspect of machine intelligence. Beyond machine learning, 
they are other capabilities like understanding language, reasoning and planning
that make up AI. 

\section{What is the opportunity?}

In the last decade, exponential improvements in computing, data and algorithms
have helped propel AI forward. Computer performance has
been doubling every two years. Data, which is powering machine
learning, has also been doubling every two years. And even 
the performance of many algorithms has been improving exponentially. 
For instance, error rates on one of AI's standard benchmarks,
the ImageNet database have been
halving every two years. Funding of AI 
has also been increasing exponentially. 
In the public sector, the UK has an \pounds 1 billion AI plan,
France has an \euro 1.5 billion AI plan, 
and Germany has an \euro 3 billion AI plan.
The European Commission has
called for \euro 20 billion in investment in AI R\&D
from public and private sources. The US and China
are making similar sized investments. Even India has
a national AI strategy with plans to invest
around \$500 million to improve agriculture, 
healthcare, education, the environment and national
security. 

In 2017, I was tasked by the Chief Scientist of Australia at the
request of the Department of Prime Minister and Cabinet to co-chair a panel
of experts from the various learned academies
to undertake a horizon scanning exercise. We looked
a decade or so out on how AI is going to impact society. 
Our goal was 
to identify how the Australian government might respond to the opportunities
and challenges that AI offers. Our report \cite{acola} looked beyond the usual economic
concerns. It argued that AI offers an opportunity to improve
our well being: economic, environmental and societal. 
The economic opportunity is, of course, to improve productivity, reduce
labour costs, promote health and safety, etc. But AI should not be seen just for this.
It is also an opportunity to make society more
inclusive. These are the very technologies that will give
hearing to the deaf, sight to the blind, mobility to those with
disabilities, etc. 
And these are also the technologies to help us act more sustainably,
to  heat and cool homes more efficiently, to feed populations
more economically, to protect ecosystems and wildlife, etc.
AI offers Bhutan these same opportunities as it does Australia.
These are technologies to improve your economic,
societal and environmental well being. 

This is perhaps best seen in the burgeoning research area using
AI to tackle the UN's seventeen Sustainable Development Goals (SDGs).
Organizations like the ``AI for Good'' Foundation are applying
AI to  help solve social, economic, and environmental problems. 
The Foundations promotes a common vision for the AI research community,
provides fora in which researchers, practitioners, policy-makers, and the
public can come together, and 
offers incentives and funding for AI research
outside traditional areas like defense and advertising
and towards socially beneficial ends.  

\section{How do you win the AI race?}

AI is often described as a race. And with giants like the US and China
investing billions of dollars in winning that race, it's hard to
imagine that a small country like Australia, or an even smaller
one like Bhutan has any chance in winning. Indeed, Kai-Fu Lee, 
an AI expert and former executive at Apple, Microsoft and Google 
has suggested that Europe isn't even in the race for third place
\cite{kaifulee}. It is, however, wrong to think of this as a race
with a single winner.
There is, of course, a scientific race to build AI. And as with
any scientific race, only the first get the credit. But the race to 
apply and develop AI is one that the whole planet can win.

A good analogy is the race one hundred or so years ago
to develop and apply electricity. It is true that people like Edison
and Westinghouse received the initial patents. And it was their 
companies that initially gained many of the economic benefits. 
But today, electricity is used by companies and individuals 
around the world. It is all of us that share the benefit. AI will
be similar. It will be a pervasive technology like electricity: in
every home, office and factory. And all of can share the 
benefits. Finland's AI plan is a fine case in point. As a small
country, Finland doesn't have the resources to compete
with the US and China and win the scientific race to 
build AI. But the Finnish AI plan sets out the more realistic
goal of being a winner in the race to apply AI \cite{finland}. 

\section{What are the risks?}

It would be irresponsible not to identify the risks that AI pose
alongside the opportunities. These risks break down into
economic, political and societal. 

\begin{description}
\item[Economic risks:]
Concerns about the economic impact of AI have
often focused on issues like employment and inequality 
in developed nations. Will robots take over many 
middle-class jobs? Will AI increase the inequality that
is already fracturing these countries? There has been less analysis
on the economic risks for developing nations. For example,
will automation replace the outsourcing of work to developing
nations? Will AI and other important and topical issues like climate 
change have a disproportionate impact on developing
nations? More work is needed to understand the risks. 
However, it is already clear that doing nothing is not
an option. AI is coming, and developing and developed nations
alike need to prepare economically. 
\item[Political risks:]
Democratic institutions around the world are 
under siege. Trust in politicians and the political
process is at an all time low. And technologies like 
AI appear to be amplifying these changes. We see AI,
for instance, being used to micro-target voters as well as
to generate deepfakes of audio and video that
never took place. Technology companies
are unsurprisingly under increasing pressure to address 
misinformation and polarization on their platforms. 
\item[Societal risks:]
Novelists like Orwell and Huxely have painted
pictures of dystopian futures that it is becoming
apparent technologies like AI could take us towards.
AI driven face recognition software
can, for example, be used to surveil and oppress 
populations on an unprecedented scale. AI 
could impact on many other human rights,
even the right to life if we see AI being used on
the battlefield to decide who lives and who dies. 
Even when restricted to large technology
companies, we see AI driving societal outcomes that are
highly undesirable such as filter bubbles and radicalism. 
\end{description}
 
\section{What can a pebble do?}

No nation can sit on its hands and ignore the coming
changes. There are perhaps half a dozen actions that
a nation like Bhutan could take today that would 
ensure the AI revolution improves the well being 
of all of its citizens. Indeed, with the guidance and trust
in institutions like the Royal Family, Bhutan is arguably
better placed than many to take advantage of these
changes. 

\begin{description}
\item[Education:] Finland is aiming to teach 20\% of its
population (1 million out of a population of just over 5 million)
about the basics of AI with an online course, ``Elements of AI''. 
In its first year, they have already reached 1\% of the population. If Bhutan
wants its citizens to profit from the coming AI revolution, it 
should have a similar ambition to equip them with the necessary
skills. This could, for instance, be included as 
part of Gyalsung (National Service). In addition to 
lifting skills broadly, strategic investment in 
(overseas) postgraduate scholarships could help provide 
valuable technical expertise to government. It is not well known that there are
only around 10,000 PhDs in AI worldwide. A modest
investment could therefore have significant returns. 
\item[Government:] Perhaps the institution best
placed to profit from AI is government. It has the challenging
job of delivering services efficiently and effectively to its 
citizens on a scale that is larger than any 
commercial operation. It collects more data than any almost any other institution and 
often is more trusted by citizens than any other institution. 
Applying AI to deliver better services
such as health care, education and welfare
starts with data. In fact, with universal health care
and education, Bhutan is better placed than many developing
nations to collect the data that will drive better delivery of 
these services. First and foremost should be a data plan. 
After this, the Bhutanese government needs to adopt an agile, rapid 
prototype and iterate software cycle to ensure
its AI projects succeed. It may help to assemble a
``tiger team'' with strong skills in AI and outside the
conventional departmental structures. 
\item[Regulation:] Nations and supranational bodies like the
EU are starting to regulate the use of AI. 
It is not enough to let the market decide how
technologies like AI impact on our lives. We can
already see problems starting to develop. Every
nation state needs to consider where it might 
usefully regulate these technologies. For instance, 
most countries have strict rules on the use of conventional
media like TV for political purposes, especially in the run up to
elections. Social media is arguably even more persuasive, and 
yet is much less regulated. Most countries hold soap adverts on TV to higher
standards than the political messages distributed by social media. Perhaps
micro-targeting of political adverts should be banned? If you
have a political message, you can target voters just on age (are they
old enough to vote?) and location (are they in my voting district?). 
This would maintain freedom of speech, yet might help
prevent the polarization we see today. 
\item[Diplomacy:] The two nations bordering Bhutam, China and
India have ambitious plans to exploit AI. But perhaps the most
challenging area in which they plan to develop AI is on the
battlefield. Thirty nations have so far called for a pre-emptive
ban on the use of fully autonomous weapons at the United Nations. 
Bhutan, with it Buddhist traditions of non-violence, would be well
placed to take moral leadership on this issue and help 
ensure the region's security is not destabilized by the 
introduction of such weapons. The world will be a very unsafe
place if such weapons turned up on the Indian, Pakistan, Kashmir
or Chinese borders. Now is the time to raise the alarm. 
\end{description}

\section{Conclusions}

Bhutan is a little gem within South Asia. 
His Majesty King Jigme Singye, The Fourth Druk Gyalpo, The King Father
of Bhutan was far sighted in putting the improvement of the well being of 
the citizens of Bhutan at the centre of their government. In addition,
by investing in technologies like the Chukha Hydropower Project,
the King helped provide the economic wealth the achieve this goal. 
His Majesty King Jigme Khesar Namgyel Wangchuck has the 
opportunity to continue this foresighted intervention by embracing
technologies like AI that can continue to improve their well being.

\bibliographystyle{alpha}
\bibliography{/Users/tw/Documents/biblio/a-z2}

\begin{thebibliography}{FHO16}

\bibitem[APL17]{finland}
Pekka Ala-Pietil\"{a} and Ilona Lundstr\"{o}m.
\newblock Finland’s age of artificial intelligence: Turning {Finland} into a
  leading country in the application of artificial intelligence. objective and
  recommendations for measures.
\newblock Technical report, Publications of the Ministry of Economic Affairs
  and Employment, 2017.

\bibitem[FHO16]{frey2}
C.B. Frey, C.~Holmes, and M.A. Osborne.
\newblock Technology at work v2.0: The future is not what it used to be.
\newblock Technical report, Citi {GPS}: Global Perspectives \& Solution, 2016.

\bibitem[FO13]{frey}
C.B. Frey and M.A. Osborne.
\newblock The future of employment: How susceptible are jobs to
  computerisation?
\newblock Technical report, Oxford Martin School, 2013.

\bibitem[Jee20]{deepfakes}
C.~Jee.
\newblock An {Indian} politician is using deepfake technology to win new
  voters.
\newblock {\em {MIT} Technology Review}, 2020.

\bibitem[Lee18]{kaifulee}
Kai-Fu Lee.
\newblock {\em AI {Superpowers}: {China}, {Silicon} {Valley}, and the {New}
  {World} {Order}}.
\newblock Houghton Mifflin Harcourt, 2018.

\bibitem[LW19]{acola}
N.~Levy and T.~Walsh.
\newblock The effective and ethical development of artificial intelligence: An
  opportunity to improve our wellbeing.
\newblock Technical report, Australian Council of Learned Academies, 2019.

\bibitem[RV17]{pwc}
Dr. Anand~S. Rao and Gerard Verweij.
\newblock Sizing the prize: What’s the real value of {AI} for your business
  and how can you capitalise?
\newblock Technical report, PwC, 2017.

\end{thebibliography}

\end{document}